\documentclass{article}

\usepackage[preprint]{corl_2026}

\usepackage{amsmath,amssymb,amsfonts}
\usepackage{graphicx} %
\usepackage{booktabs} %
\usepackage[titletoc]{appendix} %
\usepackage{titletoc} %
\usepackage{wrapfig} %
\usepackage[subtle,tracking=normal,wordspacing=normal]{savetrees}
\usepackage{caption}

\usepackage[normalem]{ulem}

\definecolor{fBlue}{rgb}{0.122, 0.467, 0.706}  %
\definecolor{fOrange}{rgb}{0.855, 0.549, 0.220}  %
\definecolor{fGreen}{rgb}{0.173, 0.627, 0.173}  %

\newcommand{\blfootnote}[1]{%
  \begingroup
  \renewcommand{\thefootnote}{}%
  \footnotetext{#1}%
  \endgroup
}

\title{\Large Constrained Whole-Body Tracking for Humanoid Robots}

\newcommand{\authorhref}[3][black]{\href{#2}{\textcolor{#1}{#3}}}
\author{
    \bfseries
    \authorhref{https://danielpmorton.github.io/}{Daniel Morton}\textsuperscript{1},\,\,
    \authorhref{https://stanfordasl.github.io/people/pranit_mohnot/}{Pranit Mohnot}\textsuperscript{1},\,\,
    \authorhref{https://profiles.stanford.edu/marco-pavone}{Marco Pavone}\textsuperscript{1,2}\\\vspace{-5pt}\\
    \textsuperscript{1}\href{https://www.stanford.edu/}{Stanford University},
    \textsuperscript{2}\href{https://www.nvidia.com/en-us/research/}{NVIDIA Research}
    \vspace{-15pt}
}

\begin{document}
\maketitle

\vspace{-2mm}
\begin{abstract} Recent advances in reinforcement learning (RL) have demonstrated impressive whole-body agility for humanoid robots, yet ensuring safety and satisfying constraints -- particularly those specified after training -- remains a challenge. Towards this goal, we present ConstrainedMimic, a control framework that leverages whole-body kinematics and dynamics for real-time constraint enforcement within RL tracking policies. By integrating principles from operational space control and control barrier functions, we enable the satisfaction of arbitrary runtime constraints on both the kinematic reference motion and the underlying dynamics, with efficient computation at frequencies above 1~kHz. 
In whole-body motion-tracking experiments in simulation and hardware, we demonstrate collision avoidance (both with the robot body and external obstacles), joint limits, and center of mass stability constraints. By remaining consistent with the current contact mode and tracking objectives, we minimally restrict the capabilities of the policy when constraints are active. 
Project page: \mbox{{\small\url{https://stanfordasl.github.io/constrainedmimic}}}. 
\blfootnote{Correspondence to: \href{mailto:dmorton@stanford.edu}{dmorton@stanford.edu}}
\end{abstract}


\section{\label{sec:introduction}Introduction}
Modern humanoid robots are becoming increasingly capable of whole-body motion tracking, locomotion, and manipulation, with reinforcement learning (RL) policies that achieve striking agility on hardware~\cite{ze2025twist2, liao2025beyondmimic, luo2025sonic}. 
As these systems are deployed, especially when they must interact with novel scenes, other equipment, or humans, a framework to enforce previously unseen safety constraints becomes critical. 
An ideal runtime safety framework would be minimally invasive, policy-agnostic, easily configurable for arbitrary deployment-time constraints, and capable of real-time speeds (\(\geq 50\) Hz).

Recent work has led to significant advances in humanoid whole-body control, demonstrating capable tracking across teleoperation~\cite{ze2025twist2, ze2025twist, he2024omnih2o, ben2024homie}, multi-task control \cite{liao2025beyondmimic, he2025hover}, and expressive behaviors at scale \cite{luo2025sonic}. Despite this diversity, these works share a common architectural paradigm: reference kinematic trajectories from retargeted motion capture systems, virtual-reality headsets, or high-level planners are passed to an RL motion tracking policy, which takes these (often dynamically infeasible) references and outputs joint commands that produce closely-tracking dynamically-feasible trajectories on the real robot.
This separation provides two natural points at which safety can be considered without modifying the tracking policy: the input kinematic reference, and the output dynamics. 

With these desiderata and problem structure in mind, control barrier functions (CBFs) may be a natural choice to embed safety into humanoid whole-body control. Commonly used in optimization-based control, CBFs enforce safety via forward-invariance of a set defined by a differentiable barrier function \(h(\mathbf{z})\), and are typically applied as a safety filter on top of a nominal, unsafe controller~\cite{CBFTheoryAndApplications}. While CBFs for humanoids have been explored in prior work~\cite{hsu2015bipedal, nguyen2016stepping, khazoom2022selfcollision, paredes2024safe, yang2025shield, yang2025cbf}, safety for whole-body control remains underexplored, particularly in the context of learning-based motion tracking policies.

\begin{figure*}[t!]
    \centering
    \includegraphics[width=0.9\linewidth]{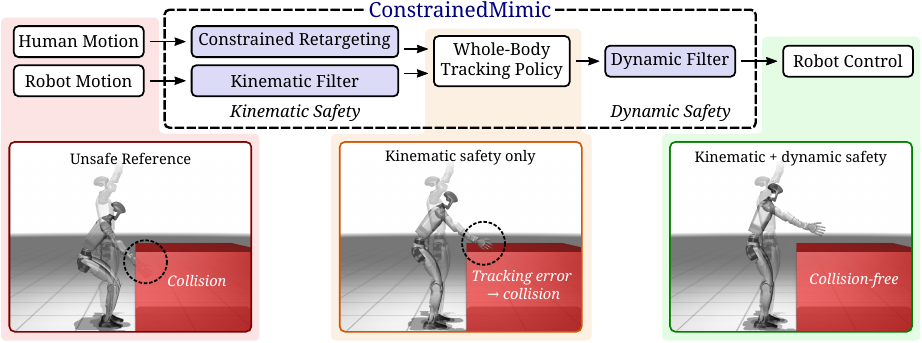}
    \caption{\textbf{\textit{Where does safety fit into a learning-based humanoid motion tracking stack?}} We approach safety at the kinematics and dynamics levels, addressing safety from both sides (input and output) of the policy. 
    On the kinematics side, constraints can naturally fit into the IK-based retargeting between human and robot form-factors, or be applied as a safety filter on motions already mapped to the robot geometry. On the dynamics side, a potentially unsafe command from the policy can be passed through a safety filter prior to execution on the robot.
    }
    \label{fig:block_diagram}
    \vspace{-2mm}
\end{figure*}

\subsection{\label{sec:related_work}Related Work}
\textbf{Whole-Body Control of Humanoid Robots.} For years, the dominant approach to whole-body control was purely model-based, including operational space control~\cite{park_khatib_2006_contact, jorda2022constraint, sentis2007thesis} and broader optimization-based methods~\cite{kuindersma2016optimization, wensing2024optimization}. More recently, RL-based whole-body control has become a leading paradigm, with systems demonstrating whole-body teleoperation~\cite{ze2025twist2, ze2025twist, he2024omnih2o, ben2024homie}, operation across multiple control modes~\cite{he2025hover}, diffusion-based skill composition~\cite{liao2025beyondmimic}, and motion tracking at foundation-model scale~\cite{luo2025sonic}. In this setting, reference trajectories are produced by retargeting human motion to robot kinematics via inverse kinematics (IK) and optimization~\cite{joao2025gmr, yang2025omniretarget, kim2025pyroki, Zakka_Mink_Python_inverse_2026}, in both online and offline contexts. Broadly, these recent RL-based methods take advantage of modeling structure in robot kinematics as an interface for control, while relying on the expressiveness of learning-based policies for robustness and recovery behaviors.

\textbf{Learning-Based Humanoid Safety.} Recent works in safety for learning-based humanoid control have explored several directions. First, methods that prioritize compliance~\cite{lu2025gentlehumanoid, chen_cao_2025chip, margolis2025softmimic} limit risk during incidental contact with humans, the environment, or the robot itself. Also critical to safety is understanding when and how to stop the robot~\cite{prism2026}, and avoiding damage during falls~\cite{meng2026safefall, strauch2025robot}. However, these contexts differ from state constraint enforcement, which typically seeks to avoid undesired contact or avoid exceeding other physical limits. Towards this end, recent work compiles and compares common safety methods \mbox{\cite{sun2025spark}}, incorporates potential fields during policy training~\cite{xue2026collisionfreehumanoidtraversalcluttered}, and analyzes constraint infeasibility in cluttered settings~\cite{chen2025dexterous}.

\textbf{CBFs for Humanoid Safety.} CBFs have been applied to legged systems since early bipedal-walking demonstrations \cite{hsu2015bipedal, nguyen2016stepping}. For humanoids, acceleration-level CBFs have been used as constraints in optimization-based whole-body controllers, for self-collision avoidance in locomotion~\cite{khazoom2022selfcollision} and task-space constraints in model-based control~\cite{paredes2024safe}. However, both works enforce the CBF with a reduced-order double-integrator model, which fails to properly consider the true constrained dynamics of the system. For more recent work combining CBFs and RL, two works are closest to ours. SHIELD~\cite{yang2025shield} learns a stochastic residual on top of a planar single-integrator nominal model and enforces CBFs for safety in expectation. CBF-RL~\cite{yang2025cbf} integrates CBFs into the RL training loop for lower-body locomotion, demonstrating planar obstacle avoidance and stair-stepping. Both focus on \textit{locomotion}, rather than whole-body tracking.

\subsection{Contributions}

We make several contributions towards safe learning-based whole-body tracking. First, we identify where safety can be enforced in the motion-tracking stack (Fig.~\ref{fig:block_diagram}): at the input (kinematic reference) and the output (low-level joint commands) of the policy, and propose three methods for constraint enforcement: \textit{constrained retargeting}, \textit{kinematic filters}, and \textit{dynamic filters}. 
For each of these methods, we construct CBFs on the full \textit{contact-constrained} kinematics and dynamics, with \textit{task-consistent} filtering objectives that respect the implicit task structure of motion tracking and loco-manipulation. Together, these form \textbf{ConstrainedMimic}: a runtime safety framework for learned whole-body motion-tracking policies which requires no retraining of the underlying policy, and is capable of running well above policy frequency even on edge compute. In both simulation and hardware, we validate performance on collision avoidance, joint limits, and stability constraints, demonstrating successful constraint enforcement even on fast-moving motions.

\section{\label{sec:method}Constrained Whole-Body Tracking}

Within an RL-based motion tracking stack, constraints can be enforced at three key points, depending on the source of the reference motion: (1) in the quadratic programming (QP)-based human-to-robot kinematic retargeting pipeline, whether online or offline (Sec. \ref{sec:constrained_retargeting}), (2) as a kinematic safety layer on top of an already-retargeted motion (Sec. \ref{sec:kinematic_filters}), and (3) as a dynamic safety layer at the torque level (Sec. \ref{sec:dynamic_filters}). In this section, we discuss each approach, and their commonalities: particularly, how the \textit{contact-constrained} kinematics and dynamics influence their construction, and how to design the filter objectives to be \textit{consistent} with typical tasks such as locomotion and manipulation.

\subsection{Safety Constraints}

\textit{Refer to the Appendix, Sec.~\ref{sec:humanoid_kin_dyn} for derivations of the kinematics and dynamics terms used in this paper.}

During operation, \textit{safety} can imply a variety of constraints to impose on the system. Most commonly, this includes collision avoidance, workspace limits, joint limits, singularity avoidance, and stability constraints. For each of these, we can define a barrier function \(h\) with respect to the state \(\mathbf{z}\) as a differentiable metric of the current safety of the system, and incorporate this into an optimization-based controller or safety filter through CBF constraints. Specifically, for a control-affine system \(\dot{\mathbf{z}} = f(\mathbf{z}) + g(\mathbf{z}) \mathbf{u}\), the CBF condition requires
\begin{equation}
    L_f h(\mathbf{z}) + L_g h(\mathbf{z}) \mathbf{u} \geq -\alpha\left( h(\mathbf{z}) \right)
    \label{eq:cbf_body}
\end{equation}
where \(L_f h(\mathbf{z}) = \frac{dh}{d\mathbf{z}}(\mathbf{z}) f(\mathbf{z})\) and \(L_g h(\mathbf{z}) = \frac{dh}{d\mathbf{z}}(\mathbf{z}) g(\mathbf{z})\) are the Lie derivatives of \(h\) along the dynamics, and \(\alpha\) is an extended class \(\mathcal{K}_{\infty}\) function. Satisfying this condition renders the safe set \(\{\mathbf{z} : h(\mathbf{z}) \geq 0\}\) forward-invariant~\cite{CBFTheoryAndApplications}. Further details on CBFs can be found in the Appendix, Sec.~\ref{sec:CBFs}.

In this work, we employ the following (vector-valued) barrier functions in our experiments. Note that for kinematic safety methods, the state \(\mathbf{z}\) is simply the generalized coordinates \(\mathbf{q}\), whereas for dynamic safety methods, the state includes the generalized velocities, i.e., \(\mathbf{z} = [\mathbf{q}, \dot{\mathbf{q}}]\).

\textit{Joint limits}: Defined as the joint-space distances to the positional limits,
\begin{equation}
    h(\mathbf{z}) =
    \begin{bmatrix}
        \mathbf{q} - \mathbf{q}_{\text{min}}\\
        \mathbf{q}_{\text{max}} - \mathbf{q} \\
    \end{bmatrix}
    \label{eq:joint_limits_cbf}
\end{equation}
\textit{Collision avoidance}: Defined as the distance to collision between two bodies \((i, j)\) in a set of collision pairs \(\mathcal{P}\) with radii \(r_i\), \(r_j\). In this paper, we restrict our focus to sphere-sphere, sphere-plane, and sphere-cylinder collision models for simplicity, where 
\begin{equation}
    h(\mathbf{z}) = \left[\|\mathbf{x}_{i}(\mathbf{q}) - \mathbf{x}_{j}(\mathbf{q})\|_2 - r_{i} - r_{j} \right]_{(i, j) \in \mathcal{P}}
    \label{eq:collision_cbf}
\end{equation}
\textit{Center of mass (CoM) stability}: Defined as the distance between the XY projection of the robot's CoM and the edges of the feet's support polygon. We represent the support polygon in halfspace form where \(\mathbf{Ax} \leq \mathbf{b}\) for a point \(\mathbf{x} \in \mathbb{R}^2\), computed by considering the convex hull of the four contact points on each foot, where
\begin{equation}
    h(\mathbf{z}) = \mathbf{b} - \mathbf{A} \mathbf{x}_{\text{CoM}}(\mathbf{q})
    \label{eq:com_cbf}
\end{equation}
Constructing these CBF constraints requires differentiating through the barrier function \(h\) with respect to the state variable \(\mathbf{z}\) (Eq. \ref{eq:cbf_body}). To do so, we compute all terms related to the robot kinematics and dynamics with \texttt{frax} \cite{morton2026frax}, and differentiate through these with \texttt{cbfpy} \cite{morton2025oscbf}.

\subsection{\label{sec:constrained_retargeting}Whole-Body Kinematically-Constrained Retargeting}

During retargeting, we consider the following control-affine model of the \textit{fully-actuated} system subject to contact constraints (Eq. \ref{eq:constrained_kinematics}), where \(\mathbf{z} = \mathbf{q}\) and \(\mathbf{u} = \dot{\mathbf{q}}\):
\begin{equation}
    \dot{\mathbf{z}} = f(\mathbf{z}) + g(\mathbf{z}) \mathbf{u} = \mathbf{0} + \mathbf{N}_c \dot{\mathbf{q}}
    \label{eq:control_affine_constrained_kin}
\end{equation}
Here, \(\mathbf{N}_c\) is the contact null-space projection matrix, and across the following sections, the subscript \(|c\) denotes a contact-constrained quantity (projected via \(\mathbf{N}_c\)).
For each frame of interest (with an associated Jacobian \(\mathbf{J}_i\) and positive-definite weighting matrix \(\mathbf{W}_i\)), we compute the task-space twist \(\boldsymbol{\nu}_i\) to reduce the error dynamics between the frame on the robot and the desired frame on the reference human kinematics.
For retargeting between human frames and the Unitree G1, we use a weighted combination of \(n_t = \) 14 pairwise tasks with independent control over the position and orientation weighting for each. 

Given a barrier function \(h(\mathbf{z})\) and an extended class \(\mathcal{K}_{\infty}\) function \(\alpha\), we then construct the CBF constraint (Eq. \ref{eq:cbf_body}) and enforce this in the following differential inverse kinematics problem:
\begin{equation}
    \begin{aligned}
        \underset{\mathbf{u}}{\text{minimize}} \quad & \sum_{i=1}^{n_t} \Vert \mathbf{W}_i  ( \mathbf{J}_{i|c} \mathbf{u} - \boldsymbol{\nu}_i ) \Vert_2^{2} \\
        \text{subject to} \quad & L_f h(\mathbf{z}) + L_g h(\mathbf{z}) \mathbf{u} \geq -\alpha\left( h(\mathbf{z}) \right)
    \end{aligned}
    \label{eq:constrained_retargeting}
\end{equation}

We also enforce box inequality constraints on the joint velocities (\(\dot{\mathbf{q}}_{\text{min}} \leq \dot{\mathbf{q}}\leq \dot{\mathbf{q}}_{\text{max}}\)), add an \(\ell_1\) relaxation for persistent feasibility (and \textit{soft} safety guarantees) (Eq. \ref{eq:relaxed_cbf}) \cite{morton2026elastiqp}, and add additional pre/post-processing steps for improved retargeting, elaborated in the Appendix, Sec.~\ref{sec:more_retargeting_details}. When the retargeter cannot be modified (e.g.,\ for offline-retargeted motions), the analogous constraints can instead be enforced as a filter on the retargeted reference, presented next.

\subsection{\label{sec:kinematic_filters}Whole-Body Kinematic Safety Filters}

For a motion which is already defined on the robot kinematics (e.g., post-offline-retargeting), we can still enforce safety at the kinematic level without explicitly re-solving the retargeting problem with constraints in the QP. In this case, we can construct a kinematic safety filter to compute a minimally-invasive modification to the reference motion to maintain constraint enforcement.

For whole-body locomotion and manipulation (including teleoperation), we adopt a hierarchical objective structure, where the primary task is motion tracking in operational space for any end-effectors (hands and feet) not currently in contact, and the CoM. As a secondary task, we consider posture tracking in the null space of motion tracking, to fully define the remaining degrees of freedom (DoFs) of the system. Note that these tasks are \textit{consistent} with the current contact mode, projected into the null space of the contact Jacobian. Given this structure, the filtering objective is:
\begin{equation}
    \| \mathbf{W}_t (\boldsymbol{\nu} - \boldsymbol{\nu}_{\text{nom}}) \|_2^2 + \| \mathbf{W}_n (\dot{\mathbf{q}}_{N} - \dot{\mathbf{q}}_{N,\text{nom}}) \|_2^2
    \label{eq:kinematic_filter_objective_1}
\end{equation}
Writing Eq. \ref{eq:kinematic_filter_objective_1} in terms of \(\dot{\mathbf{q}}\), we can then pose the CBF-QP as:
\begin{equation}
    \begin{aligned}
        \underset{\dot{\mathbf{q}}}{\text{minimize}} \quad & 
        \left\| 
        \begin{bmatrix}
            \mathbf{W}_t \mathbf{J}_{t|c} \\
            \mathbf{W}_n \mathbf{N}_{t|c}
        \end{bmatrix}
         (\dot{\mathbf{q}} - \dot{\mathbf{q}}_{\text{nom}})
         \right\|_2^2
        \\
        \text{subject to} \quad & 
        L_f h(\mathbf{z}) + L_g h(\mathbf{z}) \mathbf{u} \geq -\alpha\left( h(\mathbf{z}) \right) %
    \end{aligned}
    \label{eq:kinematic_filter_cbf}
\end{equation}
As in Sec. \ref{sec:constrained_retargeting}, we use the same constrained control-affine kinematic model (Eq. \ref{eq:control_affine_constrained_kin}) when constructing the CBF constraints, and we pair the QP with joint velocity limits and an \(\ell_1\) relaxation. 

Both kinematic methods enforce safety only on the policy's \textit{input}; next, we consider \textit{output} safety.

\subsection{\label{sec:dynamic_filters}Whole-Body Dynamic Safety Filters}

After computing the nominal (potentially unsafe) joint PD targets from the RL tracking policy, we enforce a final layer of safety before sending the command to the robot. In contrast to Secs. \ref{sec:constrained_retargeting} and \ref{sec:kinematic_filters}, this filtering \textit{can not} be applied at the kinematic level, as the joint PD targets implicitly produce a set of actuator torques, and thus we must analyze safety at a lower level.
In this case, we employ the following control-affine model of the contact-constrained and underactuated system dynamics, obtained by applying the constrained dynamics (Eq. \ref{eq:constrained_dynamics}) and actuation model (Eq. \ref{eq:underactuation_relation}):
\begin{equation}
    \dot{\mathbf{z}} = f(\mathbf{z}) + g(\mathbf{z}) \mathbf{u} = 
    \begin{bmatrix}
        \dot{\mathbf{q}} \\
        \mathbf{M}^{-1} \left( -\mathbf{N}_c^T (\mathbf{c} + \mathbf{g}) + \mathbf{J}_c^T \boldsymbol{\mu}_{c,\dot{J}} \right)\\
    \end{bmatrix}
     + 
    \begin{bmatrix} 
        \mathbf{0} \\
        \mathbf{M}^{-1} \mathbf{N}_c^T \mathbf{S}^T  \\
    \end{bmatrix}
    \boldsymbol{\Gamma}_\text{body}
\end{equation}
Here, the state \(\mathbf{z} \) contains both the generalized coordinates and velocities \(\left[ \mathbf{q}, \dot{\mathbf{q}} \right]\), the control input \(\mathbf{u}\) is the actuated joint torques \(\boldsymbol{\Gamma}_\text{body}\), \(\mathbf{M}\) is the mass matrix, \(\mathbf{c}\) and \(\mathbf{g}\) are the Coriolis and gravity terms, \(\mathbf{J}_c\) is the contact Jacobian, \(\boldsymbol{\mu}_{c,\dot{J}}\) is the Coriolis term associated with \(\dot{\mathbf{J}}_c\), and \(\mathbf{S}\) is the actuator selection matrix (App.~\ref{sec:humanoid_kin_dyn}).

As with the kinematic filter, we consider whole-body locomotion and manipulation tasks, with a similar hierarchical objective structure. To minimally interfere with the stability of the policy, we consider the primary task to be \textit{contact force tracking} on any feet in contact with the ground, and then motion and posture tasks are posed as secondary and tertiary tasks. Motion and posture are also defined with respect to the task-space and joint-space \textit{accelerations}, rather than velocities. Given these task definitions, we can then express the objective terms of the safety filter in a weighted least-squares form for each task as
\begin{equation}
    \| \mathbf{W}_c (\mathbf{f}_c - \mathbf{f}_{c,\text{nom}}) \|_2^2
    + 
    \| \mathbf{W}_t (\dot{\boldsymbol{\nu}} - \dot{\boldsymbol{\nu}}_{\text{nom}}) \|_2^2
    +
    \| \mathbf{W}_n (\ddot{\mathbf{q}}_{N} - \ddot{\mathbf{q}}_{N,\text{nom}}) \|_2^2
    \label{eq:dynamic_filter_objective_1}
\end{equation}

Writing Eq. \ref{eq:dynamic_filter_objective_1} in terms of \(\boldsymbol{\Gamma}_{\text{body}}\), we can then pose the QP as
\begin{equation}
    \begin{aligned}
        \underset{\boldsymbol{\Gamma}_{\text{body}}}{\text{minimize}} \quad & 
        \left\|
            \begin{bmatrix}
                \mathbf{W}_c \bar{\mathbf{J}}_c^T \mathbf{S}^T \\
                \mathbf{W}_t \mathbf{J}_{t|c} \mathbf{M}^{-1} \mathbf{S}^T\\
                \mathbf{W}_n \mathbf{M}^{-1} \mathbf{N}_{t|c}^T \mathbf{S}^T
            \end{bmatrix}
            (\boldsymbol{\Gamma}_{\text{body}} - \boldsymbol{\Gamma}_{\text{body,nom}})
        \right\|_2^2
        \\
        \text{subject to} \quad & L_f h(\mathbf{z}) + L_g h(\mathbf{z}) \mathbf{u} \geq -\alpha\left( h(\mathbf{z}) \right) %
    \end{aligned}
    \label{eq:dynamic_filter_cbf}
\end{equation}
We also enforce box inequality constraints on the joint torques (\(\boldsymbol{\Gamma}_{\text{body, min}} \leq \boldsymbol{\Gamma}_{\text{body}} \leq \boldsymbol{\Gamma}_{\text{body, max}}\)), and as before, use an \(\ell_1\) relaxation for persistent feasibility.

To compute the nominal actuated torque input \(\boldsymbol{\Gamma}_{\text{body,nom}}\), we apply the actuators' PD control law, with gains \(\mathbf{K}_p\), \(\mathbf{K}_d\) and clipping to the torque limits:
\begin{equation}
    \boldsymbol{\Gamma}_{\text{body,nom}} = \text{clip} \left(\mathbf{K}_p(\mathbf{q}_{\text{body,des}} - \mathbf{q}_{\text{body}}) + \mathbf{K}_d(\dot{\mathbf{q}}_{\text{body,des}} - \dot{\mathbf{q}}_{\text{body}})\right)
\end{equation}
After solving for the safe torque input \(\boldsymbol{\Gamma}_{\text{body}}^*\) from Eq. \ref{eq:dynamic_filter_cbf}, we then map this back to a safe set of PD targets and send this to the robot:\footnote{Direct torque commands at up to 1 kHz (as on TORO \cite{toro}) would be preferable. The PD-target mapping is a practical concession to the Unitree G1's control interface and the current state of motion tracking policies.}
\begin{equation}
    \mathbf{q}_{\text{body}}^* = \mathbf{q}_{\text{body}} + \mathbf{K}_p^{-1}\left[\boldsymbol{\Gamma}_{\text{body}}^* - \mathbf{K}_d(\dot{\mathbf{q}}_{\text{body,des}} - \dot{\mathbf{q}}_{\text{body}})\right]
\end{equation}

\section{\label{sec:experiments}Experiments}

We evaluate our methods in both simulation and hardware (Unitree G1), with unmodified TWIST2~\cite{ze2025twist2} and SONIC~\cite{luo2025sonic} policies as the underlying whole-body controller. In our experiments, we address five questions, each motivating a successive piece of the framework: is safety already implicit in the policy (Sec.~\ref{sec:self_collision_experiment}); when is dynamic filtering necessary (Sec.~\ref{sec:karate_chop_experiment}); how do contact constraints shape the CBF and QP objective (Sec.~\ref{sec:contact_constraints_experiment}); where do single-step constraints fail (Sec.~\ref{sec:limbo_experiment}); and how to handle necessary contact mode changes (Sec.~\ref{sec:contact_mode_experiment}). Furthermore, we conclude with validating performance both computationally (Sec.~\ref{sec:perf}) and on hardware (Sec.~\ref{sec:hardware}).

\subsection{\label{sec:self_collision_experiment}\textit{To what extent is safety already represented in the policy?}}

\begin{figure}
    \centering
    \includegraphics[width=\linewidth]{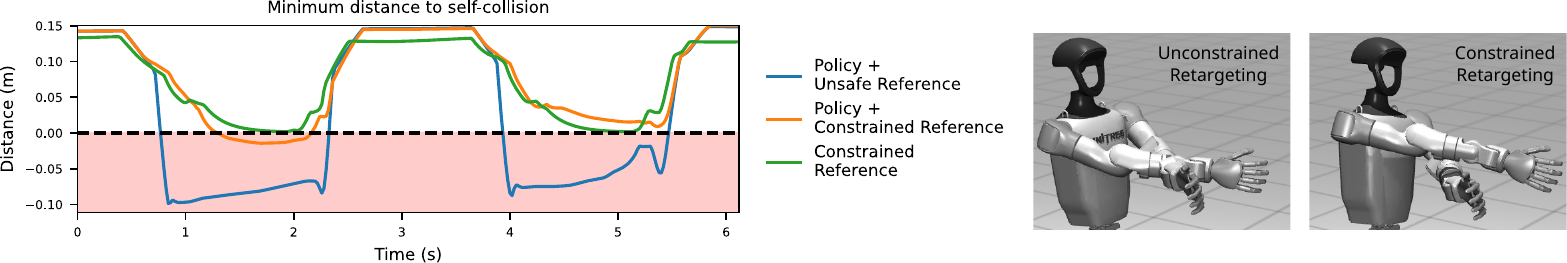}
    \caption{Self-collision constraint violation from the TWIST2 policy tracking both unfiltered and filtered kinematic references.}
    \label{fig:self_collision}
    \vspace{-3mm}
\end{figure}

Safety specifications which can be defined purely with respect to the robot model (e.g., self-collision, joint limits) are known at training time, and are often included in the reward function. However, we find that motion tracking policies such as TWIST2 do not provide any inherent guarantees of satisfying these safety constraints. In a simulated whole-body online teleoperation experiment (Fig. \ref{fig:self_collision}), passing an unsafe ``arm-crossing" reference to the policy immediately results in self-collision between the arm geometry [\textcolor{fBlue}{blue}]. Additionally, while applying a self-collision CBF during the retargeting IK process can enforce safety [\textcolor{fGreen}{green}], simply tracking this safe reference with the policy can still lead to constraint violation [\textcolor{fOrange}{orange}]. Given this, we observe that even constraints known \textit{a priori} benefit from explicit safety filtering at deployment time.

\subsection{\label{sec:karate_chop_experiment}\textit{Is dynamic safety necessary if kinematic safety is enforced?}}

\begin{figure*}[ht]
    \centering
    \includegraphics[width=0.95\linewidth]{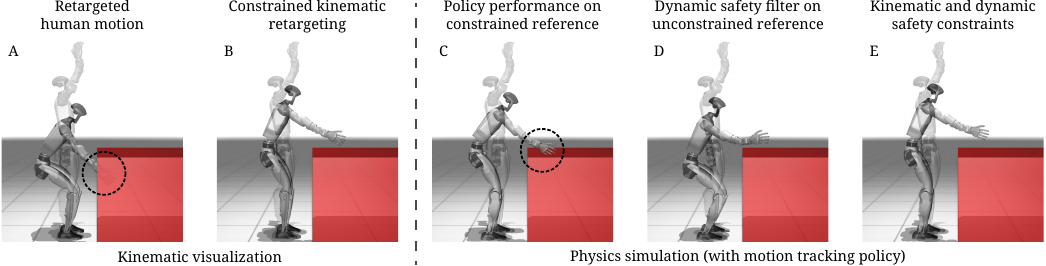}
    \caption{Handling deployment-time safety at both the kinematic and dynamics levels.}
    \label{fig:karate}
    \vspace{-3mm}
\end{figure*}

Motivated by the failure modes of purely kinematic methods in Sec. \ref{sec:self_collision_experiment}, we look at the performance of dynamic safety methods in the context of a ``karate chop" collision avoidance scenario; an external safety constraint which can \textit{not} be defined at training time. Here, we enforce a constraint (Eq. \ref{eq:collision_cbf}) between the hand and the planar surface of a table, and we find that while constrained kinematic retargeting can lead to a safe reference motion (Fig. \ref{fig:karate}B), this is no longer safe when the policy is deployed due to overshoot (Fig. \ref{fig:karate}C). When adding the dynamic safety filter (Fig. \ref{fig:karate}D--E), we see that safety is maintained regardless of the kinematic reference being safe or unsafe, but combining both kinematics- and dynamics-based filters can lead to marginally higher conservatism (Fig. \ref{fig:karate}E).

\subsection{\label{sec:contact_constraints_experiment}\textit{How do contact constraints impact the design of the safety filter?}}

\begin{figure*}[ht]
    \centering
    \includegraphics[width=0.95\linewidth]{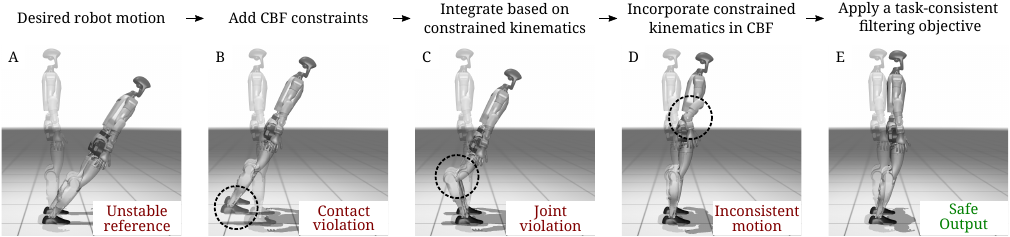}
    \caption{Ablation on the components of contact-constrained kinematic safety filters.}
    \label{fig:smooth_criminal}
    \vspace{-3mm}
\end{figure*}

Consider an \textit{antigravity lean} dance move [Jackson, 1988]; impossible to execute for a human or robot without specialized shoe fixtures or harnesses. The primary safety condition violated is center of mass (CoM) stability: the projection of the CoM into the XY plane departs from the interior of the feet's support polygon, indicating that the pose is not statically stable.\footnote{In short, the robot (referred to as Annie) would \textit{not} be OK if this were performed without a safety filter.}

Enforcing a standard min-norm safety filter (Eq. \ref{eq:cbf}) with both a CoM stability CBF (Eq. \ref{eq:com_cbf}) and joint limits CBF (Eq. \ref{eq:joint_limits_cbf}) on the full unconstrained kinematics of the system leads to the behavior shown in Fig. \ref{fig:smooth_criminal}B. The CoM remains within the original (static) support polygon, and joint limits are respected, but integrating without considering the constrained kinematics (Eq. \ref{eq:constrained_kinematics}) leads to infeasible motion where the feet leave the ground. Accounting for this, in Fig. \ref{fig:smooth_criminal}C, we now observe a violation of the CBF constraints, due to a mismatch between the kinematic model used in the CBF (unconstrained) and the true kinematic model (constrained). Lastly, in Fig. \ref{fig:smooth_criminal}D, we see that without a task-consistent filtering objective (Fig. \ref{fig:smooth_criminal}E), the filter can induce undesired motion at the boundary of safety.

\subsection{\label{sec:limbo_experiment}\textit{Where are the failure modes of short-horizon constraint enforcement?}}

\begin{figure*}[ht]
    \centering
    \includegraphics[width=0.85\linewidth]{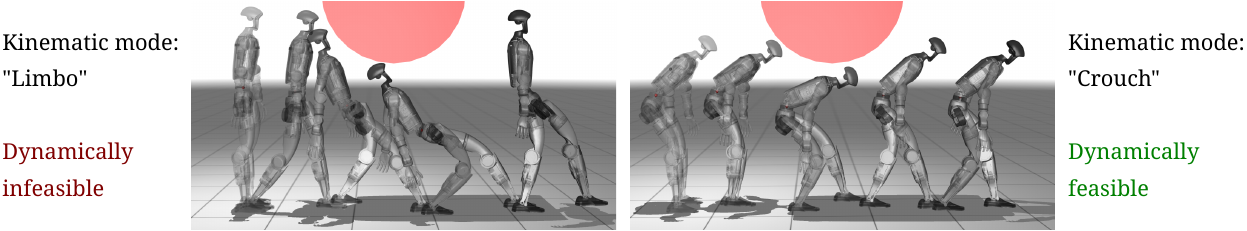}
    \caption{Impact of short-horizon safety constraints on dynamic feasibility across discrete modes.}
    \label{fig:limbo}
    \vspace{-3mm}
\end{figure*}

Consider a robot walking under a cylindrical obstacle at head-height (Fig. \ref{fig:limbo}), with collision avoidance enforced via CBF constraints on a spherized model of the robot geometry (Eq. \ref{eq:collision_cbf}). Walking safely under the obstacle can take one of two modes: ``limbo"; bending backward to avoid a collision, or ``crouch"; bending forward. These discrete modes of approaching an obstacle have distinct consequences on the dynamic feasibility of the trajectory, with ``crouch" leading to a feasible trajectory whereas ``limbo" is unstable. By nature, CBFs are a \textit{single-step} method of enforcing safety, and some constraints (such as this) inherently require longer-horizon reasoning over discrete actions or modes. In teleoperation, we can assume that a human provides this higher-level reasoning (\textit{``how should the robot approach this obstacle?"}), whereas for full autonomy, this reasoning would require an additional layer in the stack; a target for future work.

\subsection{\label{sec:contact_mode_experiment}\textit{When is it necessary to break the current contact mode to maintain safety?}}

\begin{wrapfigure}{r}{0.43\textwidth}
    \centering
    \includegraphics[width=0.4\textwidth]{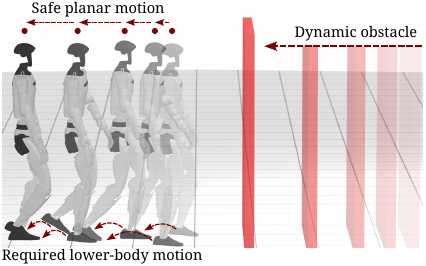}
    \caption{Tracking lower-body kinematic plans with SONIC for dynamic collision avoidance.}
    \label{fig:breaking_contact}
\end{wrapfigure}

Consider a humanoid standing at rest with a dynamic obstacle moving towards the robot (Fig. \ref{fig:breaking_contact}). In the contact-constrained kinematic setting, no motion of any feet currently in contact is permitted, leading to an infeasible problem where collision is inevitable. This safety condition inherently requires the ability to break the current contact mode and begin walking to maintain safety.

In the context of whole-body motion tracking, we cannot simply command a planar twist for a locomotion policy to track. Instead, we must explicitly define a lower-body kinematic motion to track a safe velocity, typically through a footstep planner and IK solver (similar to Eq. \ref{eq:constrained_retargeting}), or through a learned planner \cite{luo2025sonic}.

As with the ``limbo" example, this requires higher-level discrete decision making (\textit{``when should the robot take a step?"}). We propose a simple heuristic to approach this: in parallel with the whole-body CBFs, enforce an obstacle-avoidance CBF on a reduced-order model of the planar locomotion dynamics\footnote{SHIELD's residual-augmented single-integrator model would apply here \cite{yang2025shield}.}. When this constraint is active, pass the safe velocity to a lower-body kinematic planner, and enforce whole-body safety on the updated reference. This approach is practical, easy to implement, and safe in most cases, but does not give a firm guarantee of safety. Future work will explore tighter integration of discrete high-level decision making with more continuous notions of safety, and better locomotion dynamics models.

\subsection{Performance Analysis}
\label{sec:perf}

\begin{figure*}[ht]
    \centering
    \includegraphics[width=\linewidth]{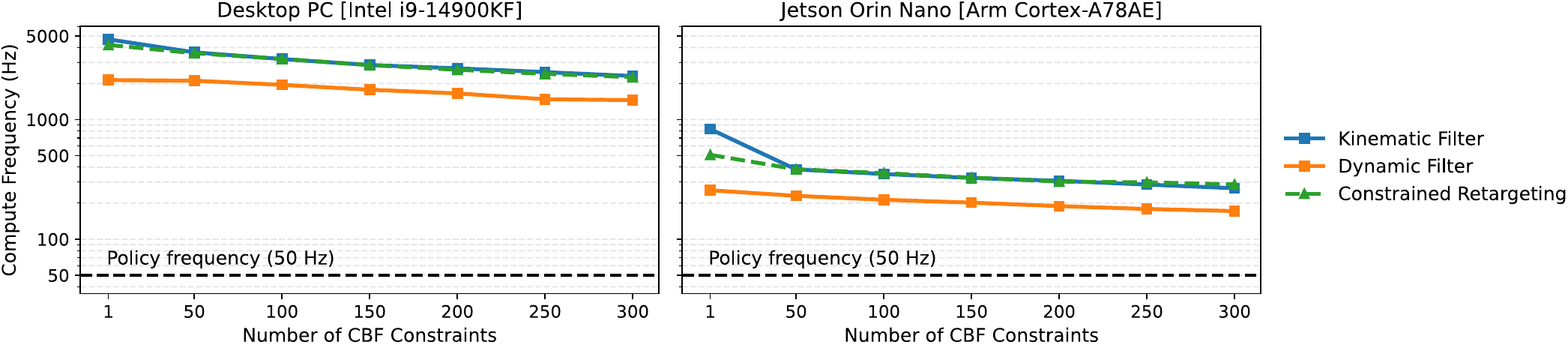}
    \caption{Compute frequencies for constrained whole-body tracking.}
    \label{fig:timing}
    \vspace{-2mm}
\end{figure*}

When implementing safety filters and inverse kinematics/dynamics methods for online control, speed is critical. Higher frequencies result in tighter feedback loops with less jitter or delay, and enable control over faster dynamics. And, for a fixed time budget, more efficient methods allow for higher scalability in the number of constraints, or the ability to run onboard on edge devices. As seen in Fig. \ref{fig:timing}, all three methods in this paper run well above the policy frequency of 50 Hz on both desktop PCs and edge devices. On desktop, even with over 300 constraints, all methods achieve \(\geq 1\) kHz control rates. For filtering, this unlocks the ability to enforce safety at higher rates than the policy, to constrain faster dynamics when new low-level state information is available (in general, the dynamic filter should be run as fast as possible given the compute budget).

In Tab. \ref{tab:constraint_numbers}, we also quantify the performance of the kinematic and dynamic safety methods on the self-collision and karate chop experiments. To evaluate these, we vary both parameters of the CBF constraint (e.g., table height for karate chop) and the playback speed of the reference motion (uniformly distributed between \(0.5 \times\) and \(2.0 \times\) the nominal rate) across 100 trials for each method and experiment. Consistently, we see that (1) the base policy is unable to maintain safety, even for conditions known at training time, (2) adding a kinematic filter significantly reduces constraint violation but does not perfectly ensure safety due to overshoot, (3) dynamic safety filters reduce constraint violation significantly due to operating at the final, lowest level of the stack, but can still have some minor violation (for reasons elaborated in the Appendix, Sec. \ref{sec:cbf_comments}), and (4) enabling constraints on both input and output levels most reliably leads to safe operation.

\begin{table}[ht]
\centering
\caption{Constraint enforcement performance on TWIST2 [\textcolor{fOrange}{orange}] and SONIC [\textcolor{fGreen}{green}] policies.}
\smallskip
\scriptsize
\begin{tabular}{@{}lcc@{\hspace{0.6em}}cc@{}}
\toprule
&
\multicolumn{2}{c}{Self-Collision} &
\multicolumn{2}{c}{Karate Chop} \\
\cmidrule(l{0.4em}r{0.4em}){2-3}
\cmidrule(l{0.4em}r{0.4em}){4-5}

& Frames in violation [\%]
& Max violation [mm]
& Frames in violation [\%]
& Max violation [mm] \\
Base Policy   & \textcolor{fOrange}{37.6 ± 0.3} / \textcolor{fGreen}{37.7 ± 0.2}                  & \textcolor{fOrange}{97.4 ± 3.0} / \textcolor{fGreen}{100.4 ± 2.7}         & \textcolor{fOrange}{20.6 ± 4.1} / \textcolor{fGreen}{18.9 ± 7.2}                  & \textcolor{fOrange}{424 ± 152} / \textcolor{fGreen}{253 ± 143}    \\
Kinematic Safety & \textcolor{fOrange}{8.5 ± 1.3} / \textcolor{fGreen}{7.5 ± 1.0}                    & \textcolor{fOrange}{12.1 ± 0.6} / \textcolor{fGreen}{52.9 ± 4.9}          & \textcolor{fOrange}{12.6 ± 8.6} / \textcolor{fGreen}{7.6 ± 8.7}                   & \textcolor{fOrange}{100.3 ± 91.3} / \textcolor{fGreen}{64.6 ± 75.7}       \\
Dynamic Safety  & \textcolor{fOrange}{0.35 ± 0.86} / \textcolor{fGreen}{1.24 ± 1.26}                    & \textcolor{fOrange}{0.50 ± 1.32} / \textcolor{fGreen}{0.43 ± 0.41}            & \textcolor{fOrange}{0.28 ± 1.16} / \textcolor{fGreen}{3.10 ± 3.57}                    & \textcolor{fOrange}{0.09 ± 0.48} / \textcolor{fGreen}{5.98 ± 8.39}            \\
Kin. + Dyn. Safety      & \textcolor{fOrange}{0.02 ± 0.10} / \textcolor{fGreen}{0.00 ± 0.00}                         & \textcolor{fOrange}{0.03 ± 0.12} / \textcolor{fGreen}{0.00 ± 0.00}                 & \textcolor{fOrange}{0.05 ± 0.35} / \textcolor{fGreen}{0.29 ± 1.10}                    & \textcolor{fOrange}{0.02 ± 0.14} / \textcolor{fGreen}{0.25 ± 1.04}            \\ \bottomrule
\end{tabular}
\label{tab:constraint_numbers}
\vspace{-2mm}
\end{table}

\subsection{Hardware Validation}
\label{sec:hardware}

\begin{figure}
    \smallskip
    \centering
    \includegraphics[width=\linewidth]{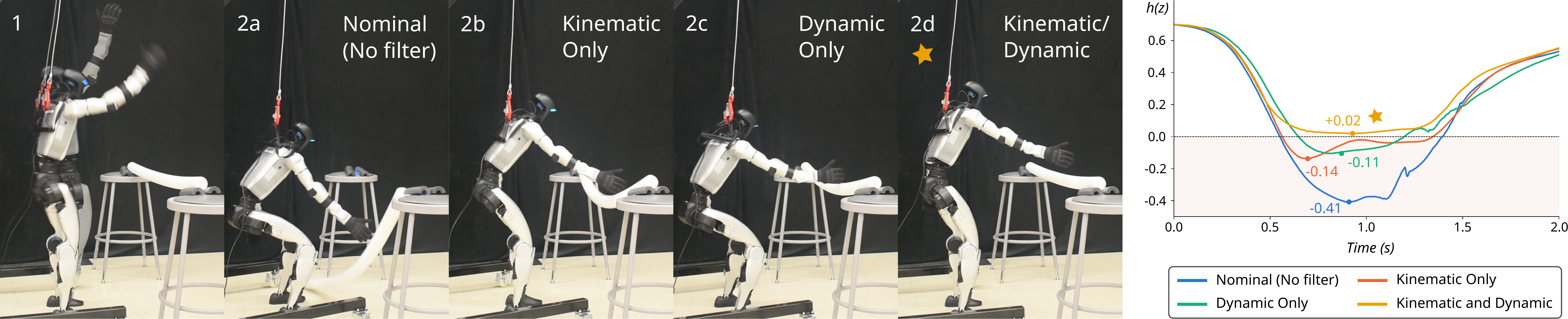}
    \caption{Hardware validation of the kinematic and dynamic safety filters on the ``karate-chop" motion of Section \ref{sec:karate_chop_experiment}. After a fast-moving initial start of the motion (1), we observe the outcomes with no filter (2a), kinematic safety only (2b), dynamic safety only (2c), and both kinematic and dynamic safety layers (2d). The plot visualizes the constraint satisfaction throughout the trajectory: \(h(\mathbf{z}) \geq 0\) when safe.
    }
    \label{fig:hardware}
    \vspace{-2mm}
\end{figure}

Building upon our simulation results, we recreate the karate-chop example from Section \ref{sec:karate_chop_experiment} and deploy this motion on the Unitree G1 hardware. Here, we use TWIST2 as the underlying general motion tracking policy, wrapped in a low-latency ROS2 interface to handle the low-level control, safety layers, and the original recording of the motion data from the Pico XR headset \cite{pico4ultra_2023}.

On hardware, we note three main differences from simulation that influence the performance of the safety filter: delays (from communication, actuators, and computation), noise (from realistic actuator measurements and state estimation), and model mismatch (unmodeled friction, imperfect inertial values). While our emphasis on speed from Section \ref{sec:perf} helps to minimize the adverse effects of delays (and noise, to some extent, with some minimal smoothing), these hardware experiments primarily seek to understand how performance varies under minor dynamics model mismatch. 

Broadly, our hardware results validate our simulation results, with one minor difference to note. Whereas in simulation (with no dynamics mismatch) the dynamic filter was the \textit{least conservative} approach that could maintain safety (Fig. \ref{fig:karate}D), on hardware, this led to small constraint violation, particularly when moving fast (Fig.~\ref{fig:hardware}, 2c and plot). Thus, this provides an even more convincing argument for why safety must be considered at \textit{both} the kinematics and dynamics levels: kinematics alone can fail due to lack of consideration of inertial effects and tracking error; dynamics alone can fail due to imperfect modeling; together these complement each other for strong constraint satisfaction.

\section{\label{sec:conclusion}Conclusion}

In this work, we have presented ConstrainedMimic: a framework for constraint enforcement in a learning-based whole-body control stack. With whole-body motion tracking policies emerging as ``foundation models" for humanoid control, ConstrainedMimic allows for enforcing arbitrary safety constraints (unknown at training time) on top of a pre-trained policy, in a minimally-invasive manner, and without any need for re-training. This combination of RL + CBFs can handle both dynamic feasibility and large numbers of nonconvex state constraints in real time, bringing model-based assurances to learning-based control in a computationally-tractable manner.

\subsection{Limitations and Future Work}

\textbf{Higher-level safety.} We focus primarily on single-step constraint enforcement, and as discussed, safety often requires higher-level logic (long-horizon or semantic reasoning) for discrete decisions.

\textbf{Safety under disturbances.} We design our safety methods to \textit{minimally interfere} with recovery behaviors in the RL policy, but do not currently account for fall-inducing disturbances.

\textbf{Uncertainty in dynamics and estimation.} Dynamic filters work well when the contact modes are well-known, but this assumes that reliable sensing or estimation is available, which is not always the case. Future work will explore integration with onboard estimation for improved generality.

\textbf{Distribution shift.} Kinematic filters (generally) result in in-distribution motions, but dynamic filters can potentially lead to out-of-distribution situations, despite a minimally-invasive design. Future work will quantify this effect and explore how fine-tuning can reduce tradeoffs in performance for safety, and will take advantage of the differentiability and GPU-compatibility of these methods.

\textbf{Onboard perception.} We analyze whole-body safety with strong assumptions about the state and geometry of collision bodies in the environment. While this is useful for initial experiments, we aim to make use of onboard perception data in the future.

\clearpage

\acknowledgments{Daniel Morton was supported by a NASA Space Technology Graduate Research Opportunity. Pranit Mohnot was supported by an NSF Graduate Research Fellowship. This work was supported in part by Generative Bionics. Thanks to Yanjie Ze and William Chong for helpful discussions throughout the project.} 

\bibliography{references}

\newpage

\appendix
\section*{Appendix}

\startcontents[sections]
\printcontents[sections]{l}{1}{\setcounter{tocdepth}{3}}
\newpage

\section{\label{sec:humanoid_kin_dyn}Background: Humanoid Kinematics and Dynamics}

\textit{Notation for this section follows from \cite{park_khatib_2006_contact, jorda2022constraint, sentis2007thesis}, and all terms are computed with \texttt{frax} \cite{morton2026frax}. For brevity, after a term is introduced, we drop the dependence on \(\mathbf{q}\) or \(\dot{\mathbf{q}}\) in further expressions.}

\subsection{Kinematics}

We consider a bipedal humanoid robot modeled as a kinematic tree with generalized coordinates \(\mathbf{q}\). These coordinates fully define the state of the robot with respect to an inertial reference frame, including both the configuration of the free-floating base \(\mathbf{q}_{\text{base}}\), and all actuated DoFs on the robot body \(\mathbf{q}_{\text{body}}\). In this paper, we model the free-floating base with six virtual joints (three prismatic and three revolute) connected to the pelvis, such that \(\mathbf{q} = \left[\mathbf{q}_{\text{base}}^T \quad \mathbf{q}_{\text{body}}^T\right]^T\).

A Jacobian \(\mathbf{J}(\mathbf{q})\) defines the relationship between an operational space velocity \(\boldsymbol{\nu}\) and the joint velocities \(\dot{\mathbf{q}}\) via \(\boldsymbol{\nu} = \mathbf{J}\dot{\mathbf{q}}\). Typically, when referring to a task \(t_i\) for a humanoid robot, we will consider \(\boldsymbol{\nu}_i\) to be a twist \(\in \mathbb{R}^6\) for end-effector pose tracking (hands, and any feet not in contact), or a velocity \(\in \mathbb{R}^3\) for center-of-mass tracking. We also denote \textit{any} inverse of a Jacobian \(\mathbf{J}\) as \(\mathbf{J}^{\#}\). Later, we will refer to \textit{dynamically-consistent} inverses via the notation \(\overline{\mathbf{J}}\), the \textit{pseudoinverse} as \(\mathbf{J}^{\dagger}\), and note that in the non-redundant full-rank case, \(\mathbf{J}^{\#} = \mathbf{J}^{-1}\).

\subsection{\label{sec:contact_kinematics}Contact Kinematics}

For any feet in contact with the world, we can define a contact Jacobian: \(\mathbf{J}_{c}(\mathbf{q})\) and its time derivative \(\dot{\mathbf{J}}_c(\mathbf{q}, \dot{\mathbf{q}})\), defining the velocity \(\boldsymbol{\nu}_c\) and acceleration \(\dot{\boldsymbol{\nu}}_c\) at the contact point. In this paper, we assume a planar contact model with a known flat ground plane, where
\begin{equation}
    \boldsymbol{\nu}_c = \mathbf{J}_c \dot{\mathbf{q}} = \mathbf{0} 
    \label{eq:no_contact_velocity}
\end{equation}
\begin{equation}
    \dot{\boldsymbol{\nu}}_c = \mathbf{J}_c\ddot{\mathbf{q}} + \dot{\mathbf{J}}_c\dot{\mathbf{q}}= \mathbf{0}
    \label{eq:no_contact_accel}
\end{equation}

\subsection{Joint Space Dynamics}

We consider a humanoid robot with the following dynamics,
\begin{equation}
    \mathbf{M} \ddot{\mathbf{q}} + \mathbf{c} + \mathbf{g} + \mathbf{J}_{c}^{T}\mathbf{f}_c = \boldsymbol{\Gamma}
    \label{eq:dynamics_with_contact_forces}
\end{equation}
where \(\mathbf{M}(\mathbf{q})\) is the joint-space mass matrix, \(\mathbf{c}(\mathbf{q}, \dot{\mathbf{q}})\) is the vector of centrifugal and Coriolis forces, \(\mathbf{g}(\mathbf{q})\) is the gravity vector, 
\(\mathbf{f}_c\) is the vector of contact forces and torques, and \(\boldsymbol{\Gamma}\) is the vector of generalized joint torques.

\subsection{Contact Space Dynamics}

At the point of contact, the projected dynamics of the system can be expressed as
\begin{equation}
    \boldsymbol{\Lambda}_c \dot{\boldsymbol{\nu}}_c + \boldsymbol{\mu}_c + \mathbf{p}_c + \mathbf{f}_c = \bar{\mathbf{J}}_{c}^{T}\boldsymbol{\Gamma}
    \label{eq:dynamics_in_contact_space}
\end{equation}
where \(\boldsymbol{\Lambda}_c(\mathbf{q})\) is the contact-space inertia matrix, 
\(\boldsymbol{\mu}_c(\mathbf{q}, \dot{\mathbf{q}})\) and \(\mathbf{p}_c(\mathbf{q})\) are the projections of the centrifugal/Coriolis forces and gravity vector into contact space, and \(\bar{\mathbf{J}}_c(\mathbf{q})\) is the dynamically-consistent generalized inverse of \(\mathbf{J}_c\). The components of the joint-space and contact-space dynamic models are related via the following:
\begin{equation}
    \boldsymbol{\Lambda}_c(\mathbf{q}) = (\mathbf{J}_c \mathbf{M}^{-1} \mathbf{J}_c^{T})^{-1}
\end{equation}
\begin{equation}
    \boldsymbol{\mu}_c(\mathbf{q}, \dot{\mathbf{q}}) = \bar{\mathbf{J}}_{c}^{T} \mathbf{c}  + \boldsymbol{\mu}_{c,\dot{J}}
    \label{eq:contact_space_mu}
\end{equation}
\begin{equation}
    \mathbf{p}_c(\mathbf{q}) = \bar{\mathbf{J}}_{c}^{T}\mathbf{g}
\end{equation}
\begin{equation}
    \bar{\mathbf{J}}_c(\mathbf{q}) = \mathbf{M}^{-1} \mathbf{J}_{c}^{T} \boldsymbol{\Lambda}_c
\end{equation}
For convenience, we define an auxiliary term \(\boldsymbol{\mu}_{c,\dot{J}}\) in Eq. \ref{eq:contact_space_mu} to describe the component of \(\boldsymbol{\mu}_{c}\) associated with \(\dot{\mathbf{J}}\), where
\begin{equation}
    \boldsymbol{\mu}_{c,\dot{J}}(\mathbf{q}, \dot{\mathbf{q}}) = -\boldsymbol{\Lambda}_c \dot{\mathbf{J}}_c \dot{\mathbf{q}}
\end{equation}

\subsection{Actuation Modeling}

The generalized torque vector \(\boldsymbol{\Gamma}\) includes components associated with both the virtual joints for the free-floating base, and the \(k\) actuated joints of the robot, i.e., \(\boldsymbol{\Gamma} = \left[\boldsymbol{\Gamma}_{\text{base}}^T \quad \boldsymbol{\Gamma}_{\text{body}}^T\right]^T\). To account for only the actuated component, we define a selection matrix \(\mathbf{S}\) such that
\begin{equation}
    \boldsymbol{\Gamma} = \mathbf{S}^{T}\boldsymbol{\Gamma}_{\text{body}}
    \label{eq:underactuation_relation}
\end{equation}
\begin{equation}
    \mathbf{S} = \begin{bmatrix}
        \mathbf{0}_{k \times 6} & \mathbf{I}_{k \times k}
    \end{bmatrix}
\end{equation}

\subsection{Constrained Kinematics}

We define the contact null space projection matrix as
\begin{equation}
    \mathbf{N}_c(\mathbf{q}) = \mathbf{I} - \overline{\mathbf{J}}_{c} \mathbf{J}_c
    \label{eq:constraint_nullspace}
\end{equation}
such that for an arbitrary input velocity vector \(\dot{\mathbf{q}}_0\), the resultant \(\dot{\mathbf{q}}\) consistent with the contact constraints is
\begin{equation}
    \dot{\mathbf{q}} = \mathbf{N}_c \dot{\mathbf{q}}_0
    \label{eq:constrained_kinematics}
\end{equation}
Note that we use \(\overline{\mathbf{J}}_c\) as the inverse when constructing \(\mathbf{N}_c\), as opposed to the pseudoinverse.

Incorporating the actuation model (Eq. \ref{eq:underactuation_relation}), the resultant velocity of the actuated joints  \(\dot{\mathbf{q}}_{\text{body}}\) is the projection 
\begin{equation}
    \dot{\mathbf{q}}_{\text{body}} = \mathbf{S} \mathbf{N}_c \dot{\mathbf{q}}_0
\end{equation}

We can then express \(\dot{\mathbf{q}}\) as a function of the actuated component \(\dot{\mathbf{q}}_{\text{body}}\) 
\begin{equation}
    \dot{\mathbf{q}} = \overline{\mathbf{S}\mathbf{N}_c}\dot{\mathbf{q}}_{\text{body}}
    \label{eq:constrained_underactuated_kinematics}
\end{equation}
where the dynamically consistent inverse of \(\mathbf{S} \mathbf{N}_c\) is
\begin{equation}
    \overline{\mathbf{S}\mathbf{N}_c}(\mathbf{q}) = \mathbf{M}^{-1}  \left(\mathbf{S}\mathbf{N}_c \right)^{T}\left(\mathbf{S}\mathbf{N}_c \mathbf{M}^{-1} \left(\mathbf{S}\mathbf{N}_c\right)^T\right)^{\dagger}
    \label{eq:S_Nc_bar}
\end{equation}
Here, we apply the pseudoinverse as \(\mathbf{S} \mathbf{N}_c\) is not necessarily full rank (which only occurs during a single-foot contact mode). In general, this implies that \(\dot{\mathbf{q}}\) can not take on arbitrary values in all cases.

\subsection{Constrained Dynamics}

Given our assumption of a fixed contact model (Eqs. \ref{eq:no_contact_velocity}, \ref{eq:no_contact_accel}), the dynamics in contact space (Eq. \ref{eq:dynamics_in_contact_space}) simplify, allowing us to solve for \(\mathbf{f}_c\) in terms of the generalized torques:
\begin{equation}
    \mathbf{f}_c = \bar{\mathbf{J}}_{c}^{T} \boldsymbol{\Gamma} - \boldsymbol{\mu}_c - \mathbf{p}_c
    \label{eq:contact_force}
\end{equation}

Substituting \(\mathbf{f}_c\) back into Eq. \ref{eq:dynamics_with_contact_forces}, we can express the constrained dynamics of the system as
\begin{equation}
    \mathbf{M} \ddot{\mathbf{q}} + \mathbf{N}_c^T \left( \mathbf{c} + \mathbf{g} \right) - \mathbf{J}_c^T \boldsymbol{\mu}_{c,\dot{J}} = \mathbf{N}_c^T \boldsymbol{\Gamma}
    \label{eq:constrained_dynamics}
\end{equation}

Eq. \ref{eq:constrained_dynamics} can equivalently be posed as a function of the actuated torques only by applying Eq. \ref{eq:underactuation_relation} and replacing \(\boldsymbol{\Gamma}\) with \(\mathbf{S}^T \boldsymbol{\Gamma}_{\text{body}}\).

\subsection{Constrained Operational Space Dynamics}

For a Jacobian \(\mathbf{J}_t\) defining a generalized velocity \(\boldsymbol{\nu}_t = \mathbf{J}_t\dot{\mathbf{q}}\) in task space, the constrained operational space dynamics take on the following form, 
\begin{equation}
    \boldsymbol{\Lambda}_{t|c} \dot{\boldsymbol{\nu}} + \boldsymbol{\mu}_{t|c} + \mathbf{p}_{t|c} = \boldsymbol{\mathcal{F}}
    \label{eq:constrained_op_space_dynamics}
\end{equation}
where \(\boldsymbol{\Lambda}_{t|c}(\mathbf{q})\) is the constrained operational space inertia matrix, \(\boldsymbol{\mu}_{t|c}(\mathbf{q}, \dot{\mathbf{q}})\) and \(\mathbf{p}_{t|c}(\mathbf{q})\) are the constrained projections of the centrifugal/Coriolis forces and gravity vector, and \(\boldsymbol{\mathcal{F}}\) is an operational space wrench.

We also define a constrained task Jacobian \(\mathbf{J}_{t|c}\),
\begin{equation}
    \mathbf{J}_{t|c}(\mathbf{q}) = \mathbf{J}_t\mathbf{N}_c
\end{equation}
which projects an arbitrary \(\dot{\mathbf{q}}_0\) into the task space while remaining in the null space of the contact constraints. Analogous to Eq. \ref{eq:constraint_nullspace}, we can also define a null space projection for the task, consistent with the contact constraints
\begin{equation}
    \mathbf{N}_{t|c}(\mathbf{q}) = \mathbf{I} - \bar{\mathbf{J}}_{t|c} \mathbf{J}_{t|c}
\end{equation}

The components of the constrained operational-space dynamics are related to the joint-space and contact-space dynamics via the following:
\begin{equation}
    \boldsymbol{\Lambda}_{t|c}(\mathbf{q}) = \left(\mathbf{J}_{t|c} \mathbf{M}^{-1} \mathbf{J}_{t|c}^{T}\right)^{-1}
\end{equation}
\begin{equation}
    \bar{\mathbf{J}}_{t|c}(\mathbf{q}) = \mathbf{M}^{-1} \mathbf{J}_{t|c}^T \boldsymbol{\Lambda}_{t|c}
\end{equation}
\begin{equation}
    \boldsymbol{\mu}_{t|c}(\mathbf{q}, \dot{\mathbf{q}}) = \bar{\mathbf{J}}_{t|c}^{T} \mathbf{c} - \left( \boldsymbol{\Lambda}_{t|c} \dot{\mathbf{J}}_{t|c} + \bar{\mathbf{J}}_{t|c}^T \mathbf{J}_c^T \boldsymbol{\Lambda}_c \dot{\mathbf{J}}_c\right)\dot{\mathbf{q}}
\end{equation}
\begin{equation}
    \mathbf{p}_{t|c}(\mathbf{q}) = \bar{\mathbf{J}}_{t|c}^{T} \mathbf{g}
\end{equation}

\section{\label{sec:CBFs}Background: Control Barrier Functions}

Consider a continuous-time dynamical system in control-affine form:
\begin{equation}
    \dot{\mathbf{z}} = f(\mathbf{z}) + g(\mathbf{z})\mathbf{u} 
\end{equation}
with state \(\mathbf{z} \in \mathcal{Z} \subseteq \mathbb{R}^{n}\), input \(\mathbf{u} \in \mathcal{U} \subseteq \mathbb{R}^m\), and locally Lipschitz continuous dynamics functions \(f : \mathbb{R}^n \rightarrow \mathbb{R}^n\) and \(g : \mathbb{R}^n \rightarrow \mathbb{R}^{n \times m}\).

Safety can be posed through the lens of set invariance. For a safe subset of the state space, \(\mathcal{C} \subset \mathcal{Z}\), if we can define a control barrier function \(h(\mathbf{z}) : \mathbb{R}^n \rightarrow \mathbb{R}\) where \(\mathcal{C}\) is the zero-superlevel set of \(h\), then a controller satisfying
\begin{equation}
    \dot{h}(\mathbf{z, u}) \geq -\alpha(h(\mathbf{z}))
    \label{eq:cbf_condition}
\end{equation}
for \(\mathbf{u} \in \mathcal{U}\) and extended class \(\mathcal{K}_{\infty}\) function \(\alpha\) will render \(\mathcal{C}\) forward-invariant \cite{CBFTheoryAndApplications}.

This condition (Eq. \ref{eq:cbf_condition}) can be integrated into a convex quadratic program (QP), paired with a min-norm objective to operate as a safety filter on a nominal (unsafe) controller:
\begin{equation}
    \begin{aligned}
        \underset{\mathbf{u}}{\text{minimize}} \quad & \| \mathbf{u} - \mathbf{u}_{\text{nom}} \|_2^2 \\
        \text{subject to} \quad & L_f h(\mathbf{z}) + L_g h(\mathbf{z}) \mathbf{u} \geq -\alpha\left( h(\mathbf{z}) \right)
    \end{aligned}
    \label{eq:cbf}
\end{equation}
where \(L_f h (\mathbf{z}) = \frac{dh}{d\mathbf{z}}(\mathbf{z}) f(\mathbf{z})\) and \(L_g h (\mathbf{z}) = \frac{dh}{d\mathbf{z}}(\mathbf{z}) g(\mathbf{z})\) are the Lie derivatives of \(h\) along the dynamics, and \(\dot{h}(\mathbf{z, u}) = L_f h(\mathbf{z}) + L_g h(\mathbf{z}) \mathbf{u}\). 

The \textit{relative degree} of a CBF refers to the number of differentiations along the dynamics required before the control input \(\mathbf{u}\) explicitly appears. CBFs require a relative degree of 1 (RD1), but with mechanical systems, CBFs are often of relative degree 2 (RD2). This high relative degree implies that \(L_g h(\mathbf{z}) = 0\), and thus, we must differentiate along the dynamics again. The second time-derivative of \(h(\mathbf{z})\) can be expressed as
\begin{equation}
    \ddot{h}(\mathbf{z}, \mathbf{u}) = L_{f}^{2}h(\mathbf{z}) + L_{g}L_{f}h(\mathbf{z})\mathbf{u}
\end{equation}
where \(L_g L_f h(\mathbf{z}) \neq 0\) for a RD2 CBF. 

Given this, we can construct a High-Order CBF (HOCBF) \cite{XiaoHOCBFs} for these RD2 constraints. Let \(h_2(\mathbf{z}) = L_fh(\mathbf{z}) + \alpha(h(\mathbf{z}))\). Then, we modify the constraint in Eq. \ref{eq:cbf} to
\begin{equation}
    L_f h_2(\mathbf{z}) + L_g h_2(\mathbf{z}) \mathbf{u} \geq -\alpha_2\left( h_2(\mathbf{z}) \right)
    \label{eq:rd2_cbf_constraint}
\end{equation}
for an additional class \(\mathcal{K}_\infty\) function \(\alpha_2\).

\subsection{\label{sec:cbf_comments}Comments on Practical CBF Implementation}

When implementing CBFs on real robots, some assumptions and conditions are typically loosened with respect to the theory. Many of these violate any formal guarantees, but despite this, CBFs can still be extremely useful tools for \textit{softer} notions of safety. Below, we note a few common conditions seen in practice:

\begin{itemize}
\item When enforcing a large number of CBFs, particularly with input constraints, these will sometimes be in conflict, resulting in an infeasible QP. In practice, relaxing the QP results in a reasonable solution that enforces (but does not guarantee) safety in most cases. Given this, we can update the QP (Eq. \ref{eq:cbf}) by introducing a slack variable,\footnote{Note: solvers such as ElastiQP \cite{morton2026elastiqp} or \texttt{qpax} \cite{TracyQPax,arrizabalaga2026differentiableinteriorpointmethodsingle} can solve this problem structure without explicit materialization of the slacks \(\mathbf{t}\) for improved performance.} \(\mathbf{t}\), to handle the constraint violation with a large penalty, \(\rho\):
\begin{equation}
    \begin{aligned}
        \underset{\mathbf{u}, \mathbf{t}}{\text{minimize}} \quad & \| \mathbf{u} - \mathbf{u}_{\text{nom}} \|_2^2  + \rho^T \mathbf{t}\\
        \text{subject to} \quad & L_f h(\mathbf{z}) + L_g h(\mathbf{z}) \mathbf{u} \geq -\alpha\left( h(\mathbf{z}) \right) - \mathbf{t} \\ 
        & \mathbf{t} \geq \mathbf{0}
    \end{aligned}
    \label{eq:relaxed_cbf}
\end{equation}
    \item The set invariance condition of CBFs is defined for continuous-time systems, but the QP is solved in discrete-time. At high sampling rates (\(\approx 1000\) Hz) this difference is negligible, but towards lower sampling rates (\(\approx 50\) Hz) this can lead to noticeable violation and/or chatter near the boundary of the safe set. For further discussion, refer to discrete-time CBFs \cite{agrawal2017discrete}.
    \item It is often assumed that the control input for the CBF constraint is unbounded, i.e. \(\mathcal{U} = \mathbb{R}^m\), but this does not hold in practice with actuator limits at the velocity and torque levels. Adding input constraints to the QP in combination with the CBF often works well in practice, but does not guarantee forward invariance of the safe set. For further discussion, see input-constrained CBFs \cite{AgrawalInputConstrainedCBFs} or the comparisons between torque-control and velocity-control CBFs with torque limits in \cite{morton2025oscbf}.
    \item Model mismatch, including imperfect actuators, miscalibrated inertial values, or unreliable contact mode estimation, can reduce the performance of the CBF when deployed on hardware.
\end{itemize}

\section{\label{sec:additional_details}Additional Implementation Details}

\subsection{\label{sec:timing}Timing and Performance}

Desktop timing values were recorded on a PC with an i9-14900KF CPU, 64 GB RAM, and an RTX 4090 GPU. Edge device timing values were recorded on a Jetson Orin Nano with an Arm Cortex-A78AE CPU, 8 GB RAM, and an integrated Ampere GPU. The Orin was run at the standard (non-Super) power mode of 15 W. \textit{All methods in this paper were run on CPU}. As per \cite{morton2026frax}, we evaluated performance with \texttt{time.perf\_counter} over 10,000 iterations, with \texttt{jax.block\_until\_ready()} called on all functions to avoid timing the asynchronous dispatch. We use JAX version 0.11.2 and the following flags: \texttt{JAX\_ENABLE\_X64=1} and \texttt{XLA\_FLAGS="--xla\_cpu\_multi\_thread\_eigen=false"}. Running single-threaded was found to give both good performance and good resource management in a ROS2 stack with many nodes.

\subsection{\label{sec:contact_modes}Contact Modes and Estimation}

As mentioned in Sec. \ref{sec:contact_kinematics}, we assume a planar contact model on the feet. For a bipedal humanoid, this yields four discrete contact modes: 0 (no contact), 1 (left foot), 2 (right foot), and 3 (both feet). All methods incorporate the current contact mode into the kinematics and dynamics models.

Regarding contact estimation, we find that at the kinematics/input level, simple heuristics can lead to fairly reliable near-ground-truth desired contact modes under minor assumptions about the motion (such as planar contact and typical walking motions). In short, we look at the relative heights and velocities of the two feet, and determine contact based on which feet are near the ground and not moving (defined by threshold parameters on the height and planar/normal velocity). These heuristics implicitly assume that mode 0 (no contact) does not occur; a reasonable assumption for typical locomotion and manipulation tasks.

For contact estimation at the dynamics/output level, currently, we evaluate the dynamic filter under well-known double-support (mode 3) contact. However, we aim to explore further integration with state estimation, such as model/optimization-based approaches \cite{kuindersma2016optimization, flayols2017estimation} or more recent work \cite{baumgartner2026cocoinekf}.

\subsection{\label{sec:more_retargeting_details}Constrained Retargeting Pipeline}

The full pipeline for computing the constrained retargeting solution is as follows. Note that some components are loosely based on \cite{joao2025gmr}, for instance, the selection of the frame correspondence between the G1 and human reference.

\textit{Pre-processing (Before the QP solve)}
\begin{enumerate}
    \item Record human reference data. We use a PICO 4 Ultra~\cite{pico4ultra_2023}, similar to~\cite{ze2025twist2}, with a custom C++ ROS2 interface for the XRoboToolkit SDK~\cite{zhao2025xrobotoolkit}. For teleoperation, high-frequency and smooth input data is critical to downstream performance, and this interface was designed to minimize latency and jitter.
    \item Adjust the desired orientations of the feet to be parallel with the floor, to better suit our planar contact model.
    \item Compute the velocity and position of the feet and incorporate this into a simple contact estimation heuristic (described in Sec. \ref{sec:contact_modes}).
    \item Rescale the positional data to approximately reflect the size difference between the human and Unitree G1.
    \item Update the heights of all bodies to put the lowest point on the feet at \(z = 0\). As previously mentioned, this assumes that mode 0 (no contact) is not considered.
\end{enumerate}
\textit{Constructing and solving the QP}
\begin{enumerate}
    \item Compute the error dynamics for the frame correspondences between the (pre-processed) human data and the current robot state.
    \item Compute the Jacobians for all frames on the robot body with \texttt{frax} \cite{morton2026frax}
    \item Compute the CBF terms with \texttt{cbfpy} \cite{morton2025oscbf}
    \item Construct the QP matrices and solve the problem with \texttt{elastiqp.jax}\footnote{Note: \texttt{qpax} \cite{TracyQPax, arrizabalaga2026differentiableinteriorpointmethodsingle} can also be used to solve the QP; while \texttt{qpax} is slower for online deployment, it is fully compatible with GPU and TPU.} \cite{morton2026elastiqp}.
\end{enumerate}
\textit{Post-processing (After the QP solve)}
\begin{enumerate}
    \item Integrate the optimal \(\dot{\mathbf{q}}\) according to the constrained kinematics
    \item Apply an exponential moving average filter to the free-floating base velocities to ensure a smooth observation
\end{enumerate}

On initialization, the first human pose in the reference motion may be quite different from the default standing pose of the robot. To align the internal state of the solver with this initial pose, we iterate until convergence in a sequential quadratic programming (SQP) fashion. Here, we perform the pre-processing step once, and then the QP solve and integration steps are executed repeatedly. Note that during this initialization phase, we integrate on the unconstrained kinematics, to allow for the necessary updates to the solver state. With a simple adaptive stepsize scheme, the SQP tends to converge in about 10 iterations.

Following convergence, we can then begin online operation, taking only a single differential IK step per timestep, and employing the constrained kinematics. We limit this to a single step, because if multiple or a variable number of differential IK steps are performed per timestep during online control (as done in some retargeters such as \cite{joao2025gmr}), adding a CBF constraint to this problem would misrepresent the \(\dot{h}(\mathbf{z})\) term.

\section{Alternate Models for Humanoid Safety}

\subsection{Humanoids on Mobile Bases}

While traditionally \textit{humanoid} implies a bipedal system, often this is also used to describe upper-torso systems on a holonomic mobile base (for instance, the Galaxea R1 robot). In this case, the mobile base and any additional elevator/torso joints can be represented as a kinematic chain of prismatic and revolute joints, attached to the upper body. This yields three main benefits over a bipedal system, in the context of modeling and constraint enforcement: (1) the model is a pure kinematic tree, with no loop closures due to contacts with the ground, (2) there are no underactuated degrees of freedom from a floating base, and (3) the system is statically stable at rest.

In this case, it generally makes more sense to use a model-based low-level controller (either inverse kinematics or dynamics), as opposed to RL. This low-level controller can easily be used in conjunction with, e.g., a learned vision-language-action model which outputs task-space commands, with the CBF enforcing safety. In this case, the \textit{task-consistent} CBF objective will reflect a hierarchy of tasks suited to the bimanual system: for instance, prioritizing tracking of the end-effectors (left and right hands) and possibly a head-mounted camera, while imposing a desired posture in the null space of the primary tasks. The QP form for the safety filter and low-level controller for this case can be found in \cite{morton2025oscbf}.

\end{document}